\documentclass[runningheads]{llncs}
\usepackage{booktabs}
\usepackage{graphicx}
\usepackage{multirow}
\begin{document}
\title{Taguchi based Design of Sequential Convolution Neural Network for Classification of Defective Fasteners\thanks{NTU-PU Collaborative Research Grant}}
%
%
\author{Manjeet Kaur\inst{1} \and
Krishan Kumar Chauhan\inst{1} \and
Tanya Aggarwal\inst{1} \and
Pushkar Bharadwaj\inst{1} \and
Renu Vig\inst{1}\and
Isibor Kennedy Ihianle\inst{2}\and
Garima Joshi\inst{1}\and
Kay Owa\inst{2}\\
}
\authorrunning{Manjeet Kaur et al.}
%
\institute{UIET, South Campus Panjab University, Sector 25, Chandigarh, 160014, India. \and
School of Science and Technology, Nottingham Trent University,\\ 50 Shakespear Street, Nottingham, NG1 4FQ, United Kingdom.\\
}
\maketitle              
\begin{abstract}
Fasteners play a critical role in securing various parts of machinery. Deformations such as dents, cracks, and scratches on the surface of fasteners are caused by material properties and incorrect handling of equipment during production processes. As a result, quality control is required to ensure safe and reliable operations. The existing defect inspection method relies on manual examination, which consumes a significant amount of time, money, and other resources; also, accuracy cannot be guaranteed due to human error. Automatic defect detection systems have proven impactful over the manual inspection technique for defect analysis. However, computational techniques such as convolutional neural networks (CNN) and deep learning-based approaches are evolutionary methods. By carefully selecting the design parameter values, the full potential of CNN can be realised. Using Taguchi based design of experiments and analysis, an attempt has been made to develop a robust automatic system in this study.  The dataset used to train the system has been created manually for M14 size nuts having two labeled classes: Defective and Non-defective. There are a total of  264 images in the dataset. The proposed sequential CNN comes up with a 96.3\% validation accuracy, 0.277 validation loss at 0.001 learning rate. 

\end{abstract}
\section{Introduction}
Bolts and nuts are common fasteners in the mechanical and automotive industries. Cold forming, hot forming, thread production, machining, hardening and tempering are the procedures that fasteners go through throughout production \cite{ref14}. Changes in the material's intrinsic characteristics, effect of vibrations, tool damage, and improper process management can result in flaws in the end products as fasteners because of the processes they go through \cite{ref4}.  Fastener raw materials might often develop cracks. Wrinkles are a form of imperfection on fasteners caused by material displacement during the forging process of nuts in particular. Deformation, dents, wrinkles, scratches, fractures, rough surface, missing and misaligned threads on the fastener surface are all faults generated by  processing \cite{ref12}. Small and medium-sized industries are the largest manufacturers of these fasteners but the process of inspection is done manually, which requires a lot of people, money, and time. Even with all of that work, the potential for errors still exists, and defective products may reach clients. As a result of the vibration, these defective fasteners may lose up, or even break in long-term \cite{ref1}. The development of automation technologies to detect bad steel fasteners could help in overcoming this challenge. The machine learning defect detection system reduced human error and effort, it makes the procedure more accurate by detecting defects that might otherwise go unnoticed by humans \cite{ref2}. The process involved calculating a collection of hand-crafted textural characteristics, which were afterward trained on a classifier \cite{ref3}. Although the automated detection process based on image processing and machine learning techniques offered great benefits it has some major drawbacks. One of the most significant is that the appearance of abnormalities changes in terms of form, size, colour, geometry, and other factors even within the same inspection work \cite{ref5}. Due to these reasons computer vision applications and deep learning based convolution neural network (CNN) can be used for better efficiency and precise detection rate. 
\section{Related Work}
Computer vision-based applications for object detection and classification using transfer learning and traditional CNNs are emerging and have become very popular in software, mechanical and electrical industrial sectors. The amount of speed, efficiency, and robustness provided by these applications has made day-to-day work simple \cite{ref8}. There are a lot of examples where the use of such deep learning models for the detection and classification of industrial fasteners has proved to be successful in achieving the goal of high accuracy in less time. Some of the examples where computer vision has exceptional contributions are taken into consideration before proposing the model. Several algorithms have been used for metal crack detection and texture feature extraction. Different edge detection operators have been used to obtain thin edge features on defective  samples \cite{ref11}. Similarly several machine learning based techniques,have been explored for classification of surface defects on rolled steel \cite{ref14}. Sharifzadeh  {\it {et al.}} applied Gaussian functions and histogram approaches to detect various flaws on steel components, with a detection rate of 88.4\% for holes, 78\% for scratches, 90.4\% for coil breaks, and 90.3\%  for rust problems \cite{ref10}. Ashour {\it {et al.}} applied support vector machine to inspect the visual texture feature from the sample data, also different kernel functions has been investigated for best performance classifier \cite{ref15}. Park {\it {et al.}}   proposed machine learning based imaging based system for defect detection for dirt, scratches, and wears on surface. CNN has been applied to check the existence of defect in the target region on an input sample image. The proposed method has been proven to be advantageous in terms of time, cost and performance as compared to manual inspection techniques \cite{ref16}. Bhandari and Deshpande proposed a modified scheme based on heuristics algorithm used by human inspectors for identifying surface imperfections to compute the features, then used SVM to classify surface images into two classes: defective and defect-free. The system was tested on a surface texture database and achieved a classification accuracy of 94.19\% \cite{ref17}. Zhao  {\it {et al.}} investigated computer vision to detect pin missing problems in transmission lines that employed bolts to link different components. Defects in these bolts can cause major problems, such as grid breakdown. To solve this problem, a CNN model was proposed, which included three key improvements for extracting small-scale bolt features and achieved a 71.4\% accuracy \cite{ref18}. A version of CNN architecture was used by  Liu  {\it {et al.}} to detect catenary support components (CSCs). It involved the integration of a detection network for CSCs utilising large scale optimized and improved Faster R-CNN with a cascade network for the detection of CSCs with small scales. The model has a good accuracy of 92.8\% \cite{ref19}. Taheritanjani {\it {et al.}}  created a system that automatically recorded, preprocessed data and compared it to a variety of supervised and unsupervised machine learning models for detecting damage in 12 different fasteners. The method also helped in determining the type of fastener used. The supervised model achieved 99\%, whereas the unsupervised model achieves 84\% \cite{ref20}. Giben  {\it {et al.}} proposed a visual inspection method for semantic segmentation and classification of material using deep convolution neural network (DCNN). In this approach they identified ten kinds of materials in total using this method, and the proposed model had a classification accuracy of 93.35\%. The detection rate for chipped and crumbling ties was 92.11\% and 86.06\%, respectively \cite{ref21}. To extract rich feature information, Kou  {\it {et al.}} used specifically built dense convolution blocks in their defect detection model, which significantly increases feature reuse, feature propagation, and the network's characterization ability. The suggested model generated 71.3\% mean Average Precision (mAP) on the GC10-DET dataset, which is publicly available, and 72.2\% mAP on the NEUDET dataset \cite{ref22}. Song {\it {et al.}} presented a CNN based technique in which they consider damage screw, surface dirt and stripped screw as defective classes, images has been taken up with industrial camera. Proposed model achieve 98\% accuracy with 1.2s of average time taken to process per image \cite{ref23}. Gai {\it {et al.}} proposed a CNN technique for detecting defects on steel surfaces in industrial parts. An industrial camera was used to collect and pre-process the data, and then the VGG architecture was used to improve network features, classification, and defect recognition capability. The proposed model was then compared to other traditional strategies and found to be more effective \cite{ref24}.
Computer vision-based applications such as object identification and classification using transfer learning or traditional CNNs have grown increasingly popular in all industries. The speed, efficiency, and robustness provided by these approaches have simplified day-to-day tasks. Researchers have employed the transfer learning approach in defect categorization but these models were trained on datasets for object detection and have no resemblance to the defect detection domain. Due to the fundamental difference between the object detection and defect detection domains, CNN models must be built from the scratch and trained on good quality dataset. Furthermore, the lack of standard datasets leads to erroneous and inconclusive findings, and there is currently no model that can be right away used for transfer learning. This work proposes a dataset of defects on the nuts and a design of sequential CNN model to handle these two key difficulties. This study also demonstrates the efficacy of the Taguchi strategy, which is a simple yet effective method for handling parameter optimization problems. The proposed method aids in determining the sequential CNN parameter value for defect classification in fasteners, according to the results. 
\vspace*{-12pt}
\begin{figure}
\centering
\includegraphics[width=0.8\textwidth]{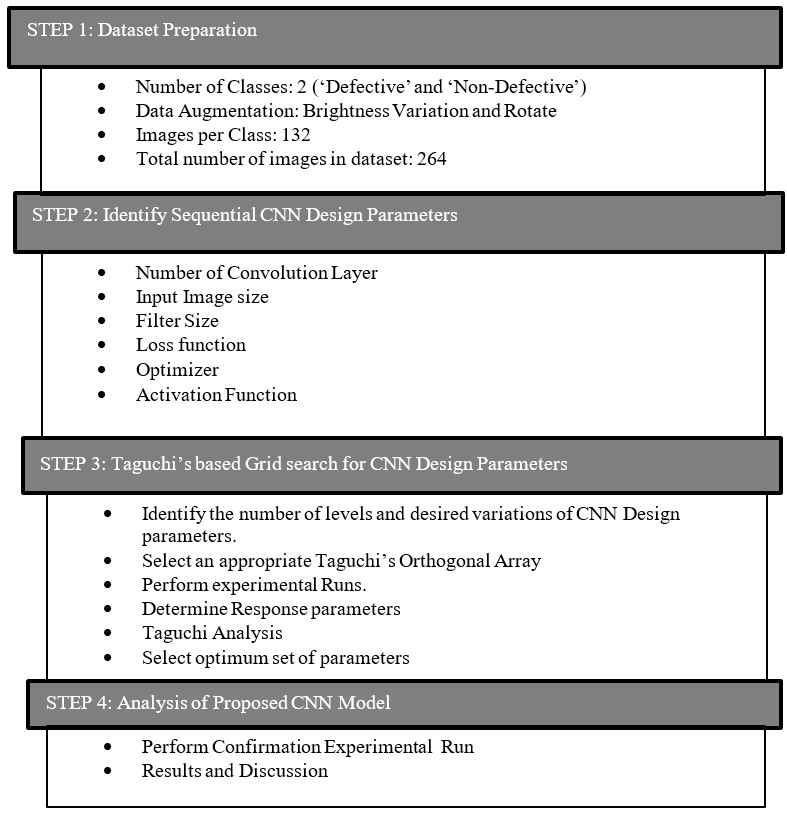}
\caption{Proposed Workflow for Taguchi based Design of Sequential Convolution Neural Network (CNN).} \label{fig1}
\end{figure}

\section{Proposed Framework for Taguchi based Design of  Sequential Convolution Neural Network (CNN) }
Fig.~\ref{fig1} shows the proposed framework for designing an optimized sequential CNN model. The details of each step are as presented below:

\subsection{Dataset Preparation} 
\begin{figure}
\centering
\includegraphics[width=0.7\textwidth]{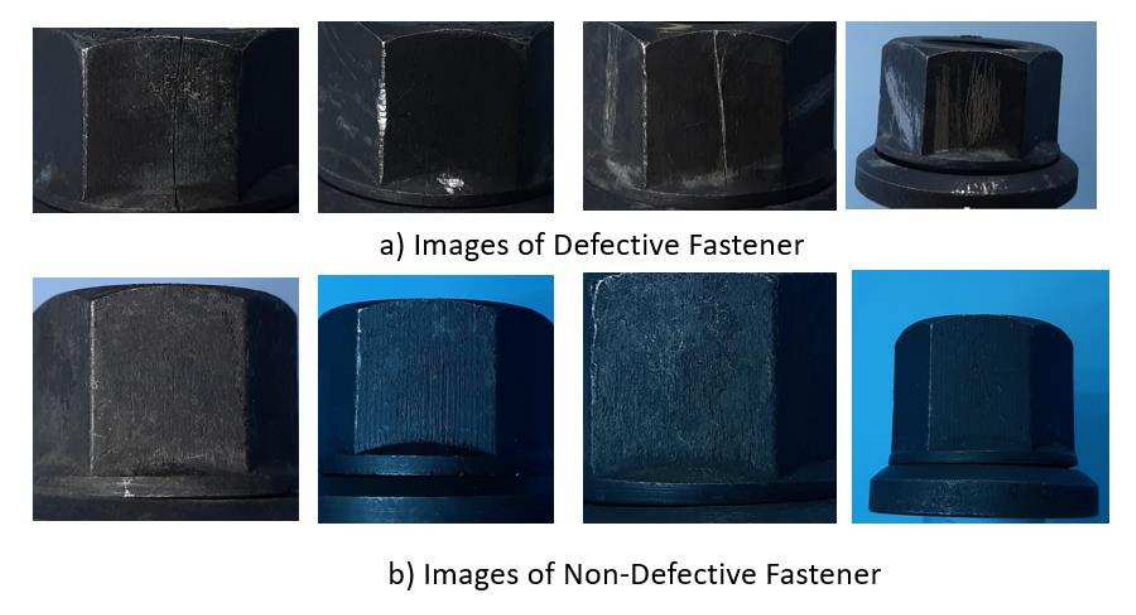}
\caption{Sample Images M14 Size Nuts in Dataset.} \label{fig2}
\end{figure}

In order prepare a practical dataset, samples of M14 size nuts with six side faces have been collected from the fastener industries. The images of size 1844*4000 pixels were captured by mobile phone camera in variable light conditions by varying the level of illumination. Image of each face of nut is captured. The real world captured dataset had the problem of class imbalance, as the number of non-defective sides of nut were more than the defective samples. To enhance the number of defective images in the dataset, augmentation in terms of scaling and brightness variation was applied. This resulted in 132 images each in ‘Defective’ and ‘Non-Defective’ class. 80\% of images were used for the training set and 20\% for the test set. Fig.~\ref{fig2}  shows the sample images from dataset. The most common types of defects are crack, dent, patches, scratches and wrinkles [6]. Here, all the defects are considered as a single class as 'Defective'.
\subsection{Sequential CNN Design Parameters} 
\begin{figure}
\centering
\includegraphics[width=0.7\textwidth]{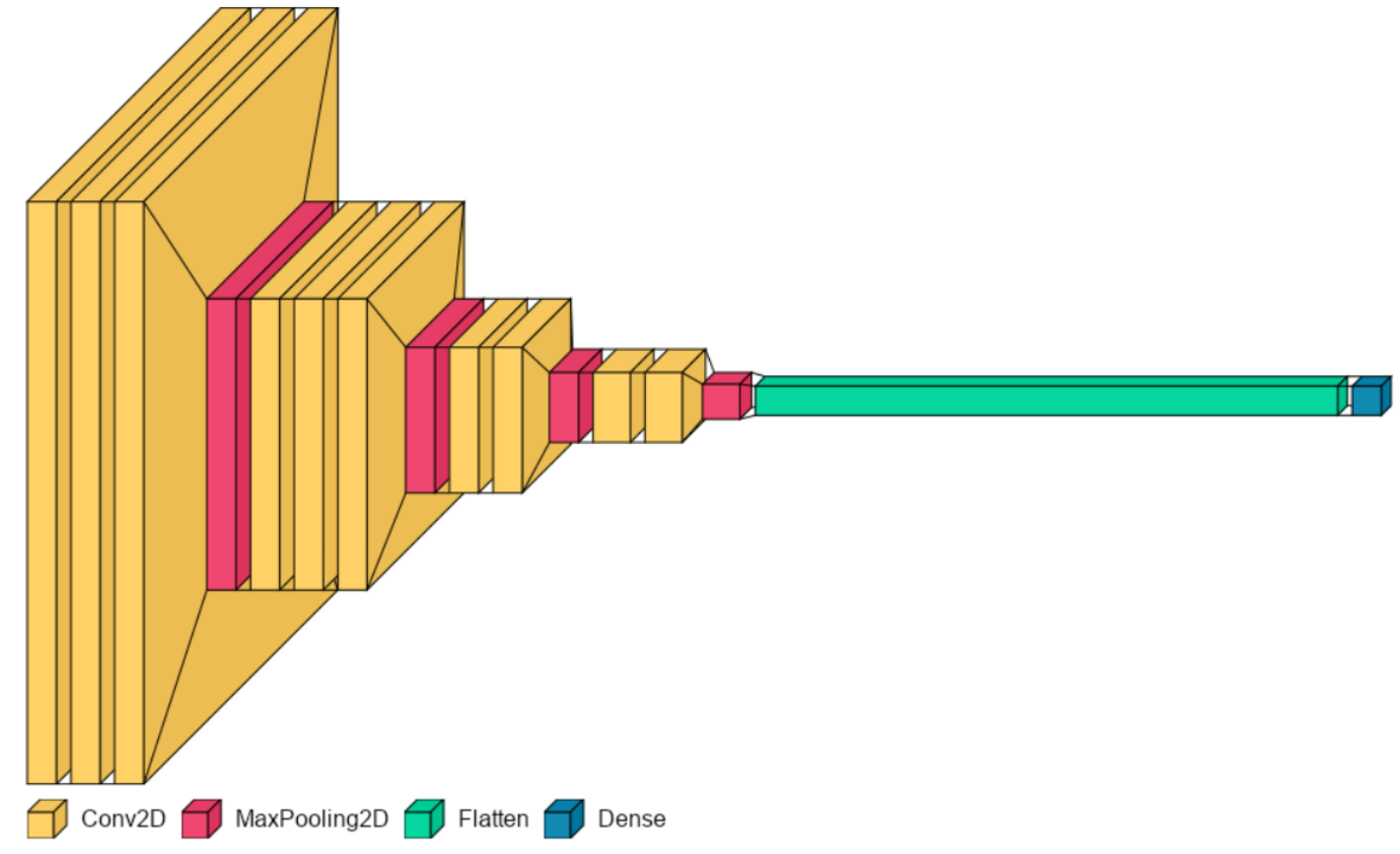}
\caption{Structure of Sequential CNN.} \label{fig3}
\end{figure}

The architecture of a basic sequential CNN is as depicted in Fig.~\ref{fig3}. These architectures can be created from scratch. To create a CNN model for a given application, early layers learn low-level characteristics, while the end-layer conducts classification based on the feature map.
\subsubsection{Number of CNN Layers}The CNN structures are motivated by the fact that increase in number of layers help in better approximation of target function and enhances the capability to learn from more detailed feature set. This makes, the number of layer as one of the essential dimension in regulating learning ability of the sequential architecture. 
\subsubsection{Input Image Size}Image size is chosen such that the fine defects such as cracks and scratches are clearly visible but larger input image size leads to more number of input is visible. However, having larger size input image leads to increase in complexity of CNN. Therefore, [100x100] and [200x200] are selected for the study. If a coloured image of 100 pixels in height and width is supplied, each pixel has one single channel, and the input layer has the form (100, 100, 3). 
\subsubsection{Optimizer} An optimizer updates the model with respect to the loss function and help in minimizing it by calculating partial derivative of the loss corresponding to weights and updating the weights in the direction opposite of the obtained gradient. This process is repeated until the loss function's minima is reached. Gradient descent takes into account the learning rate and takes larger steps if the loss is large and a smaller steps if loss is decreasing. The purpose of small learning rate is to come close to minimum value. The objective function is optimised using Stochastic Gradient Descent (SGD). It replaces the gradient with a value computed from a randomly picked data subset, this can be considered a probabilistic estimate to steepest descent. As a result, specific samples are chosen rather than the entire dataset. The global minimum can be readily attained, but when dealing with a large dataset becomes more difficult, hence SGD takes approximation of gradient descent by selecting a subset of samples. It can be particularly helpful in the case of large dataset. Adaptive Moment Estimation (ADAM) scales the learning rate using an exponentially weighted gradient. It is one of the most popular gradient descent optimizer algorithms. It keeps the record of the average of earlier gradients that are  decreasing. It is a very effective optimizer that requires very little memory and has a low training cost.
\subsubsection{Loss Function} Hinge rank loss aims to reduce the spread between the output of the model and the target vector while isolating it from all the other vectors, thus penalizing equally all the errors. Squared hinge loss computes the square of hinge loss, resulting in flat error function's surface and also making it numerically simple to handle. If a hinge loss improves performance on a binary classification issue, most probably a squared hinge loss will improve performance even more.
\subsubsection{Activation Function} Rectified Linear Unit (ReLU) is a non-linear activation function to calculate activations of convolved feature map. ReLU response maximizes beyond a certain point away from zero and has a V-shape, whereas the response is capped to the maximum value of 6 giving it a Z-shape.
\subsubsection{Filter Size}The purpose of implementing CNN is to extract key features from the input images that can characterize a class. Convolutional operation considers the local subset of pixels, therefore different levels of subsets can be explored in the image by using different filter sizes and variation in filter size capture different details. A small size filters find fine details by exploring the small regions in the image. While, large size filter finds coarse information and the model tries to find features in a large area of the input at each computation. Hence, spatial filters can be explored to improve performance with regard to learning aspect of the network. By carefully adjusting the filters, CNN can perform well both on coarse and fine details. 
In this research, an attempt is made to experiment with these basic building elements of sequential CNN in order to identify the best combination that will allow the model to perform well. The influence of each parameter adjustment and its combinations can be better understood by carefully designing the experiments.
\subsection{Taguchi based Grid search for CNN design Parameters}
\subsubsection{Taguchi based grid search} approach is relevant to all the real-world design problems which depends on number of control factors or hyperparameters. Taguchi's orthogonal array has been used in the design of experiments \cite{mont1}. It is found out to be proficient when paralleled to many other statistical designs [13]. Value of control factors must be determined in order to achieve the optimal results. Taguchi analysis aims to enhance quality as a way to deliver resilient configurations and design. The term "signal" refers to improved performance with little variance, which is referred to as "noise." The consistency of performance is a metric of robustness, which is achieved by making the design immune to the effects of uncontrollable characteristics. A two-step optimization procedure is used in Taguchi grid search. Selection of hyper parameters and desired modifications is the initial stage. Next, the orthogonal array is chosen to ensure that these control components participate equally and in a planned manner \cite{taguchi1}. The best potential performance is produced in the second stage by identifying the best arrangement of hyperparameters.
\subsubsection{Selection of Hyperparameters} Variations in object size, and distance from the camera are sources of variation (noise) in the defect categorization domain. The aim is to find the optimum hyperparameter combination that delivers high performance when subjected to fluctuations. Table 1 lists the description, value of parameters variation while performing trials. Since there is one component with four levels and five factors with two levels, L16 is the best choice among the Taguchi mixed level design possibilities for six parameters. A total of 36 trials will be required if the complete factorial design with all possible cmbinations is used. However, utilising Taguchi's orthogonal array to properly design trials, the total number of trials that must be examined for testing is reduced to 16 as indicated in Table 2.
\begin{table}[h!]
\centering
\caption{Parameters and variations.}\label{tab1}
\begin{tabular}{ccc}
\hline
{Factors  (6)} & {Name of parameter} & {Variations/Levels} \\
\hline
{A} & {No. of CNN Layers} & {  6,   8,   10,   12}\\
{B} &Image Size &[100x100], [200x200]\\
{C} & {Optimizer} & {adam,  sgd}  \\
{D} & {Loss Function} & {Hinge, Squared Hinge} \\
{E} & {Activation Function} & {ReLU, ReLU6} \\
{F} & {Filter Size} & {[}2x2{]}, {[}3x3{]} \\
\hline
\end{tabular}
\end{table}
\subsubsection{Selection of Response Parameters} In this work, the response parameters or performance metrics that are used to assess performance are test accuracy, validation accuracy, test loss and validation loss. The purpose is to maximize accuracy and minimize loss.
\section{Analysis of Results} 
\begin{figure}
\centering
\includegraphics[width=0.67\textwidth]{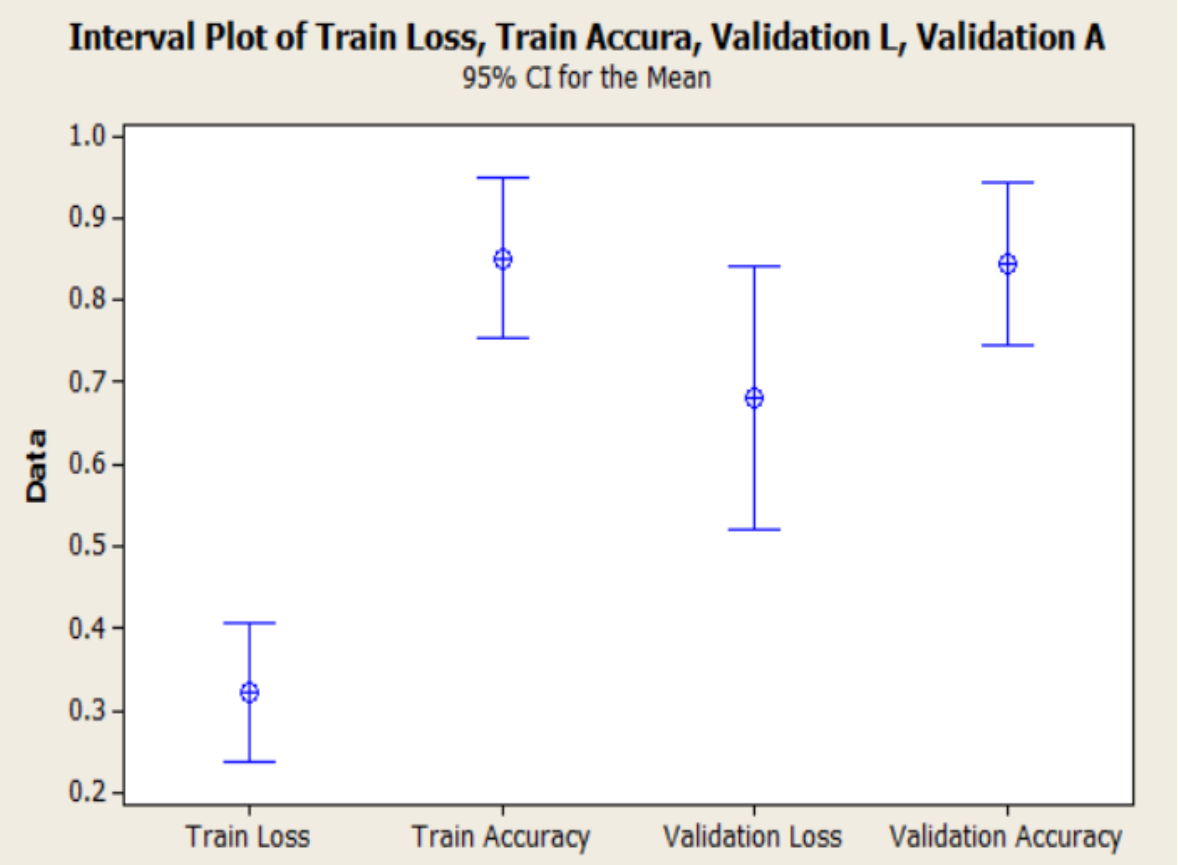}
\caption{Interval Plot.} \label{fig4}
\end{figure}

\begin{figure}
\centering
\includegraphics[width=1\textwidth]{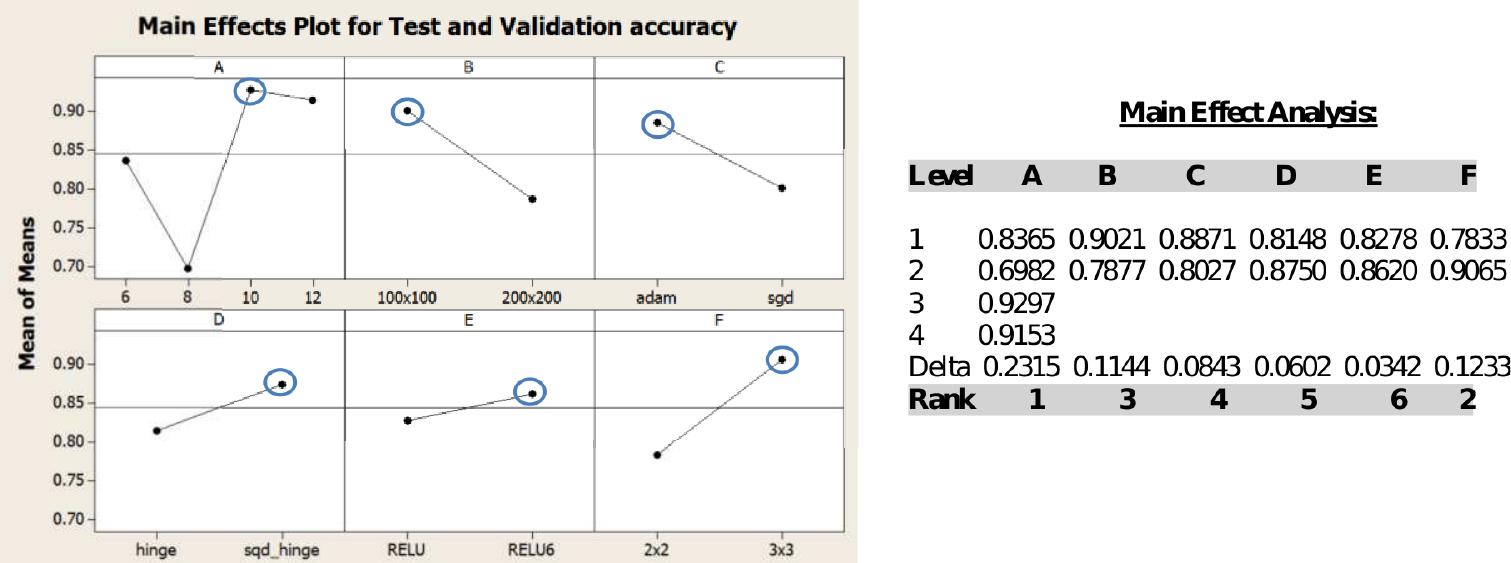}
\caption{Main Effect Plot and Main Effect Analysis from Minitab Tool.} \label{fig5}
\end{figure}

The model is implemented with Python software, and the DOE analysis is done with Minitab. The recommended set of parameters has been fine-tuned for the fastener dataset.The outcome of experiments for the possible set of combinations are recorded, each training is conducted for 500 epochs and the best result are recorded. Table 2 is analysed using Taguchi’s analysis in a Minitab tool. The highlighted text indicates the best results in Table 2.
\begin{table}
\centering
\caption{Experimental Outcome as per Taguchi's L16 Orthogonal Array.}\label{tab2}
\fontsize{10}{10}\selectfont
\resizebox{\textwidth}{!}{%
\begin{tabular}{ccccccccccc}
\hline
Exp. & \multicolumn{6}{c}{\underline{Parameter Variation}} & \multicolumn{4}{c}{\underline{Response Values}} \\
Run & \textbf{A} & \textbf{B} & \textbf{C} & \textbf{D} & \textbf{E} & \textbf{F} & \textbf{Train} & \textbf{Train} & \textbf{Val.} & \textbf{Val.} \\
& & & & & && \textbf{Loss} & \textbf{Accuracy} & \textbf{Loss} & \textbf{Accuracy} \\ \hline
1. & 6 & {[}100x100{]} & adam & Hinge & ReLU & {[}2x2{]} & 0.3885 & 0.9251 & 0.9303 & 0.9145 \\
2. & 6 & {[}100x100{]} & adam & Hinge & ReLU & {[}3x3{]} & 0.6127 & 0.8508 & 0.7954 & 0.8629 \\
3. & 6 & {[}200x200{]} & sgd & Sqd.    Hinge & ReLU6 & {[}2x2{]} & 0.3539 & 0.7552 & 0.6096 & 0.6721 \\
4. & 6 & {[}200x200{]} & sgd & Sqd.    Hinge & ReLU6 & {[}3x3{]} & 0.5114 & 0.9091 & 0.8036 & 0.8966 \\
5. & 8 & {[}100x100{]} & adam & Sqd.    Hinge & ReLU6 & {[}2x2{]} & 0.1441 & 0.8449 & 1.6285 & 0.8497 \\
6. & 8 & {[}100x100{]} & adam & Sqd.    Hinge & ReLU6 & {[}3x3{]} & 0.5757 & 0.8545 & 0.7945 & 0.8397 \\
7. & 8 & {[}200x200{]} & sgd & Hinge & ReLU & {[}2x2{]} & 0.275 & 0.1983 & 0.579 & 0.1883 \\
8. & 8 & {[}200x200{]} & sgd & Hinge & ReLU & {[}3x3{]} & 0.2206 & 0.9027 & 0.4843 & 0.9151 \\
9. & 10 & {[}100x100{]} & sgd & Hinge & ReLU6 & {[}2x2{]} & 0.3332 & 0.9455 & 0.6477 & 0.9433 \\
10. & 10 & {[}100x100{]} & sgd & Hinge & ReLU6 & {[}3x3{]} & 0.2258 & 0.9273 & 0.5374 & 0.9201 \\
11. & 10 & {[}200x200{]} & adam & Sqd.    Hinge & ReLU & {[}2x2{]} & 0.0762 & 0.8949 & 0.3033 & 0.9021 \\
12. & 10 & {[}200x200{]} & adam & Sqd.    Hinge & ReLU & {[}3x3{]} & 0.3889 & 0.9536 & 0.7223 & 0.9433 \\
13. & 12 & {[}100x100{]} & sgd & Sqd.    Hinge & ReLU & {[}2x2{]} & 0.3889 & 0.9455 & 0.6691 & 0.9433 \\
\textbf{14.} & \textbf{12} & \textbf{{[}100x100{]}} & \textbf{sgd} & \textbf{Sqd. Hinge} & \textbf{ReLU} & \textbf{{[}3x3{]}} & \textbf{0.3626} & \textbf{0.9455} & \textbf{0.4624} & \textbf{0.9433} \\
15. & 12 & {[}200x200{]} & adam & Hinge & ReLU6 & {[}2x2{]} & 0.1161 & 0.8545 & 0.5709 & 0.8528 \\
16. & 12 & {[}200x200{]} & adam & Hinge & ReLU6 & {[}3x3{]} & 0.1911 & 0.9159 & 0.3713 & 0.9216\\
\hline
\end{tabular}%
}
\end{table}
\paragraph{Interval Plot} for variation in values in train loss, train accuracy, validation loss and validation accuracy is shown in Fig.~\ref{fig4}. It is used to evaluate and compare confidence intervals for result means. An interval plot depicts a 95\% confidence interval for each group's mean. The plot indicates that the mean value of train and validation accuracy is almost equal with similar variability, while the loss values show clear displacement in terms of mean average value. The validation loss has a lot of variability and is considerably larger. Further, by analysis of outcome, efforts are undertaken to reduce loss and enhance accuracy.
\paragraph{Main Effect Plot:} Fig.~\ref{fig5} shows the Main Effect Plot for the adjustment of six hyperparameters. The main effect graphs show how each factor affects the system's overall performance. 'Main effect' is defined as when different degrees of a parameter ave varying effects on performance. When the line is horizontal, it signifies that the characteristic average is the same across all variants of that hyperparameter. If the plot is not horizontal, and changes in the factor's values affect the characteristic differently, there is a significant influence \cite{taguchi1}. The major effect plot of test and validation accuracy in Fig.~\ref{fig5} is used to determine the best set of values for each factor. The element with the greatest fluctuation has the greatest effect on the system response, and hence has Rank 1 and so on, according to the ranks of main effect analysis. The optimizer and loss function have the largest impact on system performance, while the type of activation function has the least. 
\subsection{Optimized Sequential CNN Model}
Taking into account the hyperparameters listed in Table 1, the proposed values of parameters on the basis of Taguchi analysis are selected as shown in main effect plot to maximize accuracy in Fig.~\ref{fig5}. The number of CNN layers is 10, image size is [100x100], Activation function is Relu6, loss function is squared hinge, optimizer is adam, filter size [3x3], filters in each layer are [32, 32, 32, 64, 64, 64, 128, 128, 256, 256]. Proposed sequential CNN architecture is designed and confirmation run is performed taking into consideration these parameter and summarized in Table 3. 
\begin{table}
\caption{Proposed Sequential CNN Architecture.}
\label{tab3}
\resizebox{\textwidth}{!}{%
\begin{tabular}{cccc}\\
\hline
Layer & Type & Output Shape & No. of Parameters \\
\hline
conv\_1 & 2D, Conv & (100, 100, 32) & 416 \\
conv\_2 & 2D, Conv & (100, 100, 32) & 4128 \\
conv\_3 & 2D, Conv & (100, 100, 32) & 4128 \\
max\_pooling\_1 & Pooling, max & (50, 50, 32) & 0 \\
conv\_4 & 2D, Conv & (50, 50, 64) & 18496 \\
conv\_5 & 2D, Conv & (50, 50, 64) & 36928 \\
conv\_6 & 2D,Conv & (50, 50, 64) & 36928 \\
max\_pooling\_1 & Pooling, max & (25, 25, 64) & 0 \\
conv2d\_7 & 2D, Conv  & (25, 25, 128) & 73856 \\
conv2d\_8 & 2D, Conv  & (25, 25, 128) & 147584 \\
max\_pooling\_3 & Pooling & (12, 12, 128) & 0 \\
conv2d\_9 & 2D, Conv & (12, 12, 256) & 295168 \\
conv2d\_10 & 2D, Conv & (12, 12, 256) & 590080 \\
max\_pooling\_4 & Pooling, max  & (6, 6, 256) & 0 \\
flatten\_1 & Flatten & (9216) & 0 \\
dense\_1 & Dense & (1) & 9217 \\
\hline
\multicolumn{3}{l}{Total parameters: 1,216,929} \\
\multicolumn{3}{l}{Trainable parameters: 1,216,929} \\
\multicolumn{3}{l}{Non-trainable parameters: 0} \\
\multicolumn{3}{l}{No. of CNN Layers: 10} \\
\multicolumn{3}{l}{No. of Filters in each Layer: {[}32, 32, 32, 64, 64, 64, 128, 128, 256, 256{]}} \\
\multicolumn{3}{l}{Activation Function: ReLU6} \\
\multicolumn{3}{l}{Loss Function: Squared Hinge} \\
\multicolumn{3}{l}{Optimizer: Adaptive Moment Optimizer (adam)} \\
\multicolumn{3}{l}{Batch Size: 32} \\
\multicolumn{3}{l}{Filter size in each layer: {[}3x3{]}} \\
\multicolumn{3}{l}{Classifier: Binary Support Vector Machine}\\
\hline
\end{tabular}%
}
\end{table}
\begin{figure}
\centering
\includegraphics[width=0.7\textwidth]{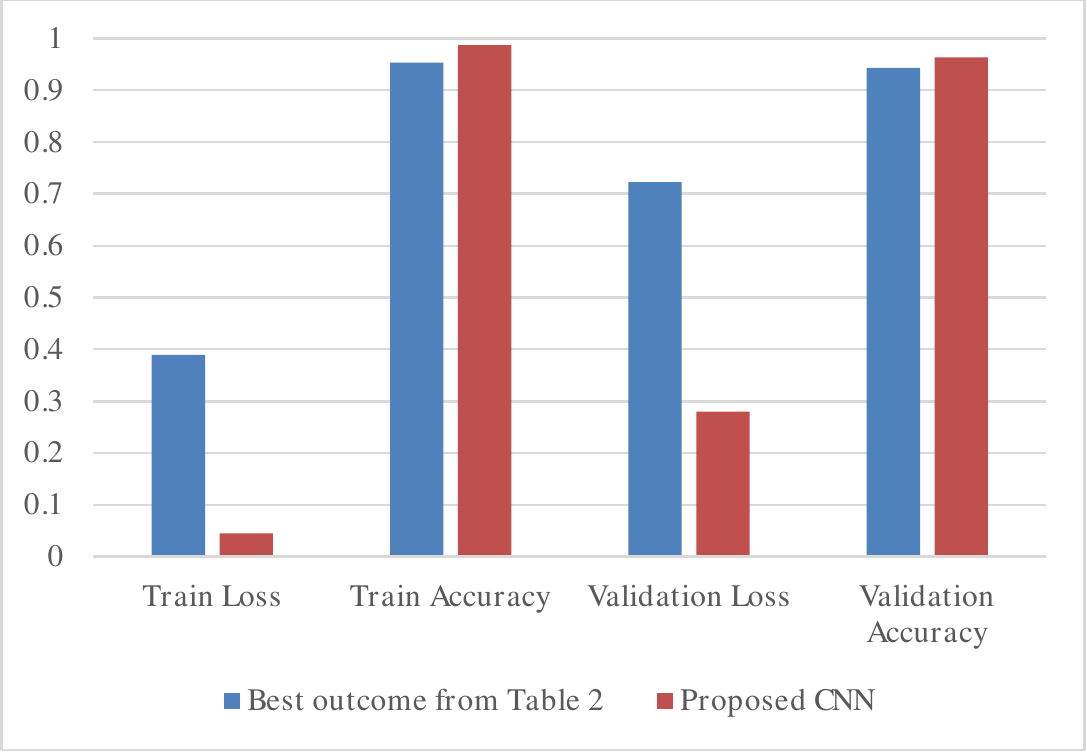}
\caption{Comparison of Results for Proposed Design.} \label{fig6}
\end{figure}
\section{Conclusion}
Automatic defect detection systems have proven to be more effective than manual inspection procedures in quality control. The goal of this work is to use Taguchi-based design of trials and analysis to create a robust autonomous system for  visual inspection. A dataset of pictures of nut surface defects is developed and used. A sequential CNN architecture is created using Taguchi design of experiments and analysis. The findings suggest that choosing the right hyperparameters is crucial for attaining low loss values and higher accuracy in a specific application. Training accuracy is 98.69\%, training loss is 0.0449, validation loss is 0.2794, and validation accuracy is 96.3\% in this study.
System resilience is seen with consistent training and validation outcomes, that is low loss and high accuracy are achieved.\\
In future work, more comprehensive study will be carried out taking into various categories of defects.  

%
%
%
\bibliographystyle{splncs04}
\bibliography{mybibliography}

\begin{thebibliography}{10}
\providecommand{\url}[1]{\texttt{#1}}
\providecommand{\urlprefix}{URL }
\providecommand{\doi}[1]{https://doi.org/#1}

\bibitem{ref15}
Ashour, M.W., Halin, A.A., Khalid, F., Abdullah, L.N., Darwish, S.H.:
  Texture-based classification of workpiece surface images using the support
  vector machine. International Journal of Software Engineering and Its
  Applications  \textbf{9}(10),  147--160 (2015)

\bibitem{ref17}
Bhandari, S.H., Deshpande, S., Deshpande, S.: A simple approach to surface
  defect detection. 2008 IEEE Region  \textbf{10},  8--10 (2008)

\bibitem{ref14}
Caleb, P., Steuer, M.: Classification of surface defects on hot rolled steel
  using adaptive learning methods. In: KES'2000. Fourth International
  Conference on Knowledge-Based Intelligent Engineering Systems and Allied
  Technologies. Proceedings (Cat. No. 00TH8516). vol.~1, pp. 103--108. IEEE
  (2000)

\bibitem{ref2}
Chen, J., Liu, Z., Wang, H., N{\'u}{\~n}ez, A., Han, Z.: Automatic defect
  detection of fasteners on the catenary support device using deep
  convolutional neural network. IEEE Transactions on Instrumentation and
  Measurement  \textbf{67}(2),  257--269 (2017)

\bibitem{ref4}
Elangovan, S., Boopathy, T., Kannan, R.: Fabrication and analysis of polymer
  bolt and nut assembly by additive manufacturing system. Journal of Emerging
  Technologies and Innovative Research (JETIR)  \textbf{6}(6) (2019)

\bibitem{ref24}
Gai, X., Ye, P., Wang, J., Wang, B.: Research on defect detection method for
  steel metal surface based on deep learning. In: 2020 IEEE 5th Information
  Technology and Mechatronics Engineering Conference (ITOEC). pp. 637--641.
  IEEE (2020)

\bibitem{ref21}
Giben, X., Patel, V.M., Chellappa, R.: Material classification and semantic
  segmentation of railway track images with deep convolutional neural networks.
  In: 2015 IEEE International Conference on Image Processing (ICIP). pp.
  621--625. IEEE (2015)

\bibitem{ref22}
Kou, X., Liu, S., Cheng, K., Qian, Y.: Development of a yolo-v3-based model for
  detecting defects on steel strip surface. Measurement  \textbf{182},  109454
  (2021)

\bibitem{ref19}
Liu, W., Liu, Z., Nunez, A., Han, Z.: Unified deep learning architecture for
  the detection of all catenary support components. IEEE Access  \textbf{8},
  17049--17059 (2020)

\bibitem{mont1}
Montgomery, D.C.: Design and analysis of experiments. John wiley \& sons (2017)

\bibitem{ref8}
Neogi, N., Mohanta, D.K., Dutta, P.K.: Review of vision-based steel surface
  inspection systems. EURASIP Journal on Image and Video Processing
  \textbf{2014}(1),  1--19 (2014)

\bibitem{taguchi1}
Oehlert, G.W.: A first course in design and analysis of experiments. Kluwer
  Academic Publishers (2010)

\bibitem{ref16}
Park, J.K., Kwon, B.K., Park, J.H., Kang, D.J.: Machine learning-based imaging
  system for surface defect inspection. International Journal of Precision
  Engineering and Manufacturing-Green Technology  \textbf{3}(3),  303--310
  (2016)

\bibitem{ref3}
Saiz, F.A., Serrano, I., Barandiar{\'a}n, I., S{\'a}nchez, J.R.: A robust and
  fast deep learning-based method for defect classification in steel surfaces.
  In: 2018 International Conference on Intelligent Systems (IS). pp. 455--460.
  IEEE (2018)

\bibitem{ref5}
Shaheen, M.A., Foster, A.S., Cunningham, L.S., Afshan, S.: Behaviour of
  stainless and high strength steel bolt assemblies at elevated
  temperatures—a review. Fire Safety Journal  \textbf{113},  102975 (2020)

\bibitem{ref10}
Sharifzadeh, M., Amirfattahi, R., Sadri, S., Alirezaee, S., Ahmadi, M.:
  Detection of steel defect using the image processing algorithms. In: The
  International Conference on Electrical Engineering. pp.~1--7. Military
  Technical College (2008)

\bibitem{ref23}
Song, L., Li, X., Yang, Y., Zhu, X., Guo, Q., Yang, H.: Detection of
  micro-defects on metal screw surfaces based on deep convolutional neural
  networks. Sensors  \textbf{18}(11), ~3709 (2018)

\bibitem{ref20}
Taheritanjani, S., Schoenfeld, R., Bruegge, B.: Automatic damage detection of
  fasteners in overhaul processes. In: 2019 IEEE 15th International Conference
  on Automation Science and Engineering (CASE). pp. 1289--1295. IEEE (2019)

\bibitem{ref11}
Xue, B., Wu, Z.: Key technologies of steel plate surface defect detection
  system based on artificial intelligence machine vision. Wireless
  Communications and Mobile Computing  \textbf{2021} (2021)

\bibitem{ref12}
Xue, P., Jiang, C., Pang, H.: Detection of various types of metal surface
  defects based on image processing. Traitement du Signal  \textbf{38}(4)
  (2021)

\bibitem{ref1}
Yun, J.P., Choi, S., Kim, J.W., Kim, S.W.: Automatic detection of cracks in raw
  steel block using gabor filter optimized by univariate dynamic encoding
  algorithm for searches (udeas). NDT and E International  \textbf{42}(5),
  389--397 (2009). \doi{10.1016/j.ndteint.2009.01.007}

\bibitem{ref18}
Zhao, Z., Qi, H., Qi, Y., Zhang, K., Zhai, Y., Zhao, W.: Detection method based
  on automatic visual shape clustering for pin-missing defect in transmission
  lines. IEEE Transactions on Instrumentation and Measurement  \textbf{69}(9),
  6080--6091 (2020)

\end{thebibliography}
\end{document}